\documentclass[10pt,twocolumn,letterpaper]{article}

\usepackage{wacv}
\usepackage{times}
\usepackage{epsfig}
\usepackage{graphicx}
\usepackage{amsmath}
\usepackage{amssymb}
\usepackage{booktabs}
\usepackage{multirow}
\usepackage[accsupp]{axessibility}

\def\wacvPaperID{1459} 

\wacvalgorithmstrack   

\wacvfinalcopy 

\ifwacvfinal
\usepackage[breaklinks=true,bookmarks=false]{hyperref}
\else
\usepackage[pagebackref=true,breaklinks=true,colorlinks,bookmarks=false]{hyperref}
\fi

\begin{document}

\title{MASTAF: A Model-Agnostic Spatio-Temporal Attention Fusion Network for Few-shot Video Classification}

\author{
Xin Liu \and 
Huanle Zhang \and
Hamed Pirsiavash \and
Xin Liu \and
University of California, Davis\\
{\tt\small \{rexliu,dtzhang,hpirsiav,xinliu\}@ucdavis.edu}
}

\maketitle
\thispagestyle{empty}

\begin{abstract}
We propose MASTAF, a Model-Agnostic Spatio-Temporal Attention Fusion network for few-shot video classification. MASTAF takes input from a general video spatial and temporal representation,e.g., using  2D CNN, 3D CNN, and Video Transformer. Then, to make the most of such representations, we use self- and cross-attention models to highlight the critical spatio-temporal region to increase the inter-class variations and decrease the intra-class variations. Last, MASTAF applies a lightweight fusion network and a nearest neighbor classifier to classify each query video. We demonstrate that MASTAF improves the state-of-the-art performance on three few-shot video classification benchmarks(UCF101, HMDB51, and  Something-Something-V2), e.g., by up to 91.6\%, 69.5\%, and 60.7\% for five-way one-shot video classification, respectively.
\end{abstract}

\section{Introduction}

Few-shot learning has received increasing attention in video classification for its potential to reduce the video annotation cost significantly~\cite{otam}.
In few-shot video classification, the video samples in the training and test sets are from different classes (i.e., unseen classes in the test set). To classify an unlabeled video sample (query), a few-shot video classification model aims to classify the query to the unseen class (support set).
Inspired by the development in few-shot image classification\cite{crosstransformers,crossattention,raghu2020rapid}, recent few-shot video classification approaches using metric-learning-based methods achieve state-of-the-art performance\cite{otam,trx}. This paper targets metric-learning-based few-shot video classification.

A metric-learning-based few-shot video learning algorithm classifies a query based on the similarity between the representation of the query video and the representation of each class in the support set. Therefore, the core to metric-learning-based few-shot video classification is to design feature extraction and representation for the support sets and the query. Many feature embedding networks have been designed for this purpose. Perrett~\cite{trx} leverages attention mechanism in temporally-ordered frames from support sets to match query frames after extracting representation for each frame with pre-trained 2D Convolutional Neural Network(2D CNN). Zhang~\cite{arn} introduces permutation-invariant pooling and self-supervised learning tasks to enhance representations after extracting from a 3D Convolutional Neural Network(3D CNN) embedding network.

In few-shot scenarios, prior efforts with a 2D CNN embedding network outperformed those with a 3D CNN embedding network~\cite{otam,trx,pal}. However, there are two considerable limitations in existing work with a 2D CNN embedding network. 
The first limitation is that a complex temporal alignment strategy between the video frames for better accuracy increases computational demand and model inference runtime. For example, Perrett~\cite{trx} achieves SOTA performance on few-shot video classification by exploring all the combinations of two and three ordered sampled frames from a video for temporal information. As the number of sampled frames from a video grows, the computational cost and inference runtime increase significantly.

The second limitation is their inability to maintain high performance when replacing a 2D CNN embedding network with other advanced video representation models such as 3D CNN~\cite{I3D,r3d,c3d,S3D-G} and Video Transformer~\cite{vivit}. With the release of large-scale video datasets, video classification models' performance based on 3D CNN and Video Transformer surpasses those with 2D CNN~\cite{vivit,I3D,would}, which means such models can generate a better representation for discrimination. Therefore, one would expect that if we replace the 2D CNN in the existing few-shot video classification models with an advanced video representation model, performance should improve. However, this did not happen. Instead, Zhu~\cite{pal} found that 3D CNN models~\cite{I3D,c3d,S3D-G} do not perform better than 2D CNN models in PAL~\cite{pal}, a SOTA 2D-CNN based few-shot video classification algorithm. The main reason is that 2D-CNN approaches rely on the frame-level similarity score and temporal alignment, which do not exist in a 3D CNN embedding network.
\begin{figure*}
\centering
\includegraphics[width=1.0\textwidth]{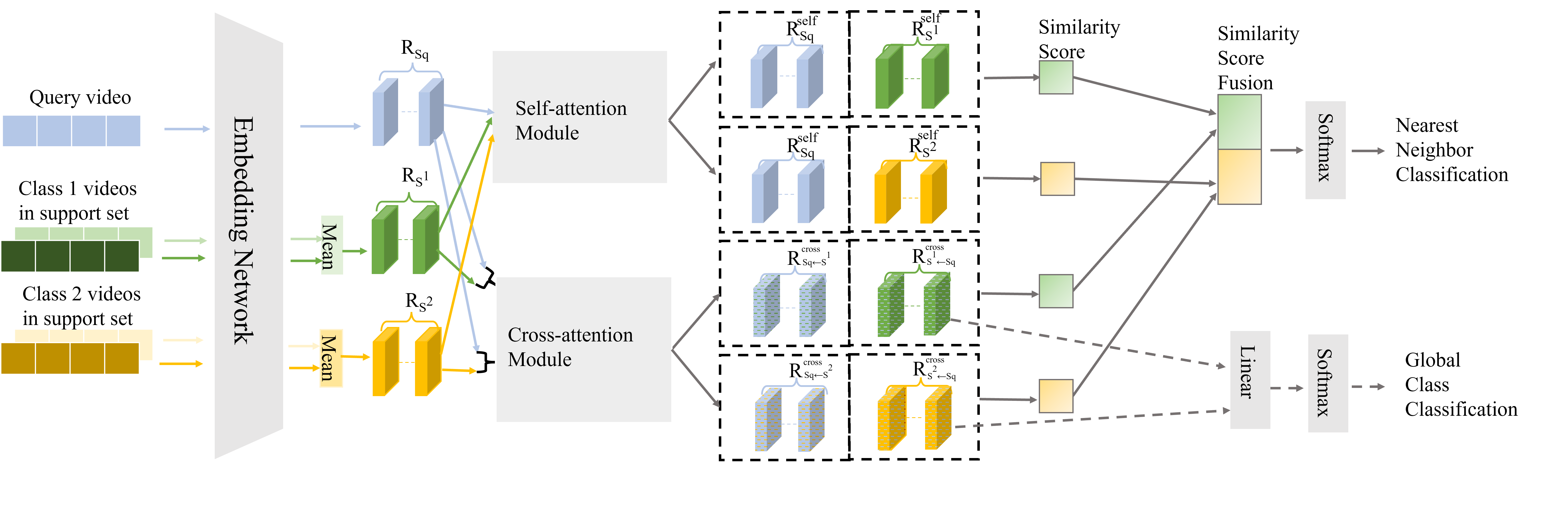}
\caption{Illustration of the Model-Agnostic Spatio-Temporal Attention Fusion(MASTAF) on a 2-way 2-shot video classification. First, we extract spatio-temporal features with a pre-trained embedding network for each video. Then, we compute a prototypical representation($R_{S^{c}}$) for each class in the support set, which is the mean of all the representations of each class. After that, we use the self-attention module to highlight spatio-temporal features for each query and support class representation and compute the similarity score of each pair of query representation and support class representation using cosine distance. In parallel, we use the cross-attention module to highlight the spatio-temporal correlation features for each pair of query representation and support class representation, and compute the similarity score using the cosine distance. The cross-attention representations of each class in the support set are fed into a global video classifier as a multi-task training set. And the fusion results of similarity scores from the self-attention module and cross-attention module are fed into the nearest neighbor classifier. Details are in Section~\ref{s:method}.}
\label{workframe}
\end{figure*}

In this paper,  we propose a model-agnostic few-shot video learning algorithm named Model-Agnostic Spatio-Temporal Attention Fusion network(MASTAF). Our key motivation is to make the most of the rapid advances in video representation learning to build a simple and efficient few-shot video learning framework. To achieve this goal, we have to address the limitations discussed above. Advanced video representation networks, such as 3D CNN and Transformer, extract spatio-temporal representations directly instead of frame-level information. To make good use of such representations, we use self- and cross-attention models to increase the weight of the critical spatio-temporal region to increase the inter-class variations and decrease the intra-class variations as shown in Figure~\ref{workframe}. The self-attention network emphasizes the regions of the representation that are essential for representing each class and the query, while the cross-attention network emphasizes the regions of the representation that enhance the discriminability between the query and the unseen classes in the support set. Then, we measure the similarity between the query and each unseen class based on the feature maps from each attention network. Last, we classify the query video by a simple yet efficient fusion network. We also add one multi-task training setting,i.e., global video classification task, to regularize the embedding module and further improve generalization performance. More details are presented in Section~\ref{s:method}

\noindent 
\textbf{Contributions} We make the following contributions. 

\noindent
1.We propose MASTAF, a simple and efficient attention-based network compatible with different video classification models for few-shot video classification. MASTAF can benefit from advanced video classification models such as 3D CNN and Video Transformer that extract good spatial-temporal representations.

\noindent
2.We design a fusion mechanism to integrate self-attention and cross-attention networks, which greatly enhances the essential spatial and temporal regions of video representation. 

\noindent 
3.We extensively evaluate MASTAF using three benchmarks, i.e., UCF101~\cite{ucf101}, HMDB51~\cite{hmdb51}, and Something-Something V2~\cite{ssv2}. Compared to the existing work, MASTAF achieves state-of-the-art performance with a 2D CNN embedding network and improves the state-of-the-art performance with a 3D CNN embedding network without additional computational cost. 
Our code is available at \url{https://anonymous.4open.science/r/STAF-30CF1}.

\section{Related work}
\noindent
\textbf{Few shot learning} Most existing few-shot learning algorithms can be divided into three categories: model-based methods~\cite{Meta-Networks,memory-modelbased}, optimization-based methods~\cite{MAML,optimizaiton}, and metric-learning-based methods~\cite{otam,trx,prototypical}. Metric-learning-based methods are more promising than other two methods in few-shot video classification since the previous work with metric-learning-based achieved better performance~\cite{otam,trx}.

Metric-learning-based method measures the distance between the representation of support samples and query samples and classifies them with the aid of the nearest neighbor to keep similar classes close and dissimilar classes far away. Particularly, Prototypcal Network~\cite{prototypical} is based on the idea that each class has a Prototypical representation which is the mean value of support set in embedding space. The few-shot learning problem then becomes the nearest neighbor in the embedding space. Our work is one of the metric-learning-based methods. We can take input from general video spatial and temporal representations extracted from different video representation models. To make the most of representations in the embedding space, we highlight the spatio-temporal features that need attention for each class while increasing the differences from other classes.

\noindent
\textbf{Few shot video classification} The first module of most metric-learning based methods for a few-shot video classification model is an embedding network that extracts features from each video. The two most commonly used approaches for embedding network are 2D CNN~\cite{otam,trx,strm,cmn,cmn-j,pal} and 3D CNN~\cite{TRAN,protogan,arn}. After using 2D CNN to extract features from each video frame, Zhu and Yang~\cite{cmn,cmn-j} introduce a memory network structure to learn optimal representations in a larger video representation space. Instead of creating a memory structure to memorize long-term information for video representation, more recent work with 2D CNN embedding networks focus on temporal alignment exploration between the query video and the support set. Cao~\cite{otam} aligns the frames between the query video and support video by temporal ordering information.
Perrett~\cite{trx} achieves SOTA on 5-way 5-shot video learning by computing the distance of temporal-relational representations between each frame of query video and support video. 
In comparison, the features extracted from the 3D CNN embedding network already contain temporal information. Therefore, recent work focus on generating general spatio-temporal video representations for unseen classes. Dwivedi~\cite{protogan} leverages GAN to generate the spatio-temporal video representations for the prototype of the unseen classes. Zhang~\cite{arn} introduces permutation-invariant pooling and self-supervised learning tasks to enhance representations whereas Bishay~\cite{TRAN} uses segment-based attention and deep metric learning. Recently, Video Transformers have become a promising option for video representation due to their long-term reasoning ability~\cite{vivit,vidir}. Although Video Transformer are not widely used in few-shot video classification, SOTA performance in video classification indicates the promise of being applied in a few-shot scenario. 

\noindent
\textbf{Attention-based learning} 
Attention mechanism enhances the learning ability of long-range dependencies in the network to highlight the critical regions of visual representations~\cite{attention}. These critical regions are useful in discriminating the differences between different classes. 
Therefore, the recent work with attention mechanisms achieve SOTA accuracy for few-shot learning tasks~\cite{trx,strm,hyrsm}. Perrett~\cite{trx} applies a cross transformer with a multi-head attention mechanism for the representation of each frame to locate the representative frames for similarity computation. Thatipelli~\cite{strm} proposes a self-attention module for the patch's representation of each frame in a video to highlight critical regions. These works adopt 2D CNN to extract the features and apply attention mechanisms for temporal alignment and frame-level feature enrichment. However, these works cannot maintain high performance when using a 3D CNN embedding network without the help from complex temporal alignment and frame-level feature enrichment. Our work is compatible with any video classification model and uses attention fusion network to highlight the spatio-temporal features, which help increase the inter-class variations and decrease the intra-class variations. Wang~\cite{hyrsm} applies multi-head self-attention for video-level feature enrichment, increasing GPU memory and computational consumption. However, our approach adopts a simple and efficient fusion layer for self-attention and cross-attention modules with a lower computational cost.

\section{MASTAF: Model-Agnostic Spatio-Temporal Attention Fusion Network}
\label{s:method}
\subsection{Problem definition}
The few-shot video classification problem aims to classify one unannotated query video into one of several annotated categories set, which we call “support set”. Each category has only a few video instances in this support set, and the model did not see these categories during the training process. 
Our paper focuses on C-way K-shot video classification, where C denotes the number of categories in the support set and K represents the number of video instances for each category in the support set. We follow the same episodic training as in the previous study~\cite{otam,trx,arn,cmn,cmn-j} that randomly select C classes with K video clips for the support set. Then we select one query video from these C classes, which is different from the K video clips in the support set. For each C-way K-shot episode, the support set contains C classes, and each class has K video clips. 

We use $S^{c}_{k}= \{f^{c}_{k,1}, f^{c}_{k,2},\dots,f^{c}_{k,n}\}$ to denote the $k^{th}$ video clip of class $c$, where $c$ belongs to $C$ and $k$ belongs to $K$, $f^{c}_{k,i}$ denotes the $i^{th}$ extracted frame from the video and $n$ denotes the total number of frames extracted from the video. For the query video, we use $S_q=\{f_1, ...,f_i,...,f_n\}$, where $f_i$ denotes the $i^{th}$ frame extracted from the query video and $n$ denotes the total number of frames extracted from the query video. The final goal is to predict $S_q$ to one of the classes.

\noindent
\subsection{The MASTAF Model}
The design principle of the MASTAF model is to highlight the critical spatio-temporal region to minimize the intra-class variations while maximizing the inter-class variations between the query video and support set. To tackle the challenge of only having few samples for the unseen class, we first extract spatio-temporal features using any video classification model. Then, we use the attention fusion module to further highlight the critical spatio-temporal region for metric learning. In parallel, we use a global classification task to regularize the embedding network. Next, we analyze each module in the MASTAF model, which is described in Figure~\ref{workframe}.

\noindent
\textbf{Embedding module} In the MASTAF model, the goal of the embedding module $f_\varphi$ is to learn the spatio-temporal representations for each video. We evenly extract frames from each video, where $n$ is the total number of frames extracted from each video. We can use any video classification model as the spatio-temporal embedding module. 

Given a frame sequence extracted from the video $S_v = \{f_1,f_2,\dots,f_n\}$, let $R_v \in \mathbb{R}^{C' \times T' \times H' \times W'}$ denote the representation learned from the embedding model:
\begin{equation}
R_v = f_\varphi(S_v). \label{feature_extract}
\end{equation}
For a video clip in the support set $S^{c}_{k}$, we use $R_{S^{c}_{k}}$ to denote the representation learned from the embedding module. We use $R_{S^{c}}$  to denote the representation of the class $c$, which is the mean of all the representations of video clips for class $c$ in the support set. And since we have only one query video in the few-shot learning task, we use $R_{S_q}$ to denote the representation for the query video clip. After we get the representations for the support set and query video, we go through two separate attention modules in parallel,.i.e, the self-attention module and the cross-attention module.
\begin{figure*}
\centering
\includegraphics[width=0.9\textwidth]{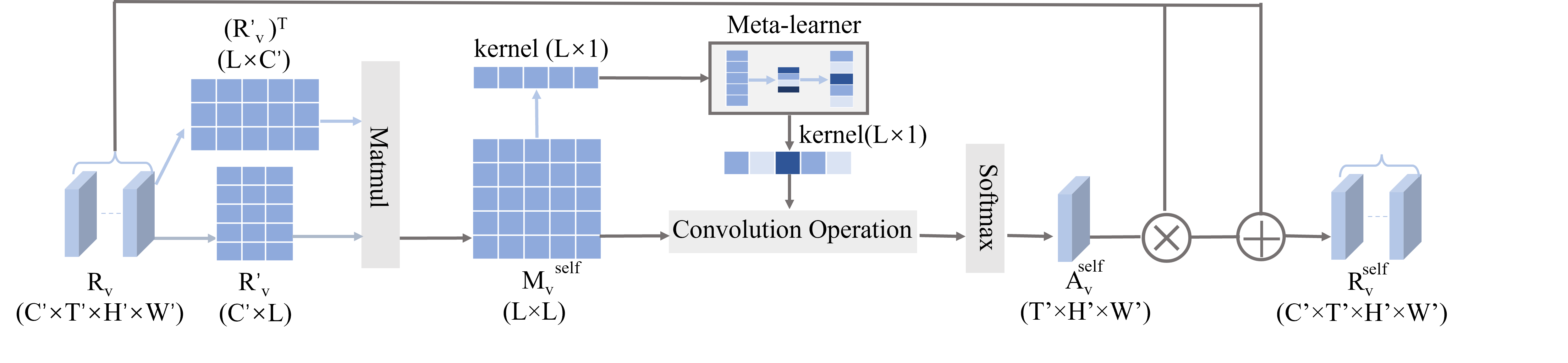}
\caption{Self-attention module}
\label{self-attention}
\end{figure*}

\noindent
\textbf{Self-attention module}
Our goal of the self-attention module is to highlight the critical information in the representation of each class. As shown in Figure~\ref{self-attention}, we first reshape each representation to $R_v^{'} \in \mathbb{R}^{C' \times L}$, where $L ( L = T' \times H' \times W' )$  is the number of spatio-temporal positions on each feature cubic map. After that, for each class in the support, $R_{S^{c}}$ becomes $R_{S^{c}}^{'}$, i.e., $[ R_1^{S^{c}},\dots R_i^{S^{c}}\dots, R_L^{S^{c}}]$, where $R_i^{S^{c}}$ denotes the feature vectors at the $i^{th}$ spatio-temporal position in the $R_{S^{c}}^{'}$. For each query video, $R_{S_q}$ becomes $R_{S_q}^{'}$, i.e., $[ R_1^{S_q},\dots R_i^{S_q}\dots, R_L^{S_q}]$, where $R_i^{S_q}$ denotes the feature vectors at the $i^{th}$ spatio-temporal position in the $R_{S_q}^{'}$. 
Then we compute the self-relation map for each representation as:
\begin{equation}
M^{self} =(R'_v)^{\intercal}R'_v, \label{self-relation}
\end{equation}
where $M^{self} \in \mathbb{R}^{L \times L}$ that denotes the self-relation map for each video, where $M_i^{self}$ denotes the self-relation at the $i^{th}$ spatio-temporal position in the feature map. Then we apply convolutional operation with a kernel $d$, i.e., $d \in \mathbb{R}^{L}$,  to fuse each position self-relation vector into an attention scalar, which is in $\mathbb{R}^{T' \times H' \times W'}$. Then we leverage a softmax function to draw self-attention for each $i^{th}$ position: 
\begin{equation}
A_i^{self} = \frac{exp((d^{\intercal}M_i^{self})/\tau)}{\sum_{j=1}^{L}exp((d^{\intercal}M_j^{self})/\tau)},
\label{self-relation-softmax}
\end{equation}
where $\tau$ is the temperature hyperparameter to amplify the variance and $A_i^{self}$ denotes the $i^{th}$ position of self-attention map $A^{self}$,i.e.,  $A^{self} \in \mathbb{R}^{T' \times H' \times W'}$.

Instead of assigning equal weight to every position, we add a meta-learner to learn the kernel $d$ dynamically to pay attention to the critical positions in the feature cubic map. First, we leverage row-wise global average pooling for $M^{self}$ to get an averaged vector $\overline{M}^{self}$,which $\overline{M}^{self} \in \mathbb{R}^{L}$. Then we use a meta-learner to learn the kernel $d$ dynamically:
\begin{equation}
d = f_{\gamma}(\sigma(f_{\delta}(\overline{M}^{self}))),
\label{meta-learner}
\end{equation}
where $f_{\delta} :\mathbb{R}^{L} \rightarrow \mathbb{R}^{l}$ and $f_{\gamma} :\mathbb{R}^{l} \rightarrow \mathbb{R}^{L}$,.i.e, $l$ denotes the scaled dimension and $\sigma$ represents the ReLU function\cite{relu}.

After we get the self-attention cubic map $A^{self}$, we leverage a residual attention
mechanism to weigh each element of the original map $R_v$ with $1+A^{self}$ to get the self-attention representation $R^{self}$ for each class:
\begin{equation}
R^{self} = R_v(1+A^{self}), 
\label{residual attention mechanism}
\end{equation}
where $R^{self} \in \mathbb{R}^{C' \times T' \times H' \times W'}$.
\begin{figure*}
\centering
\includegraphics[width=0.9\textwidth]{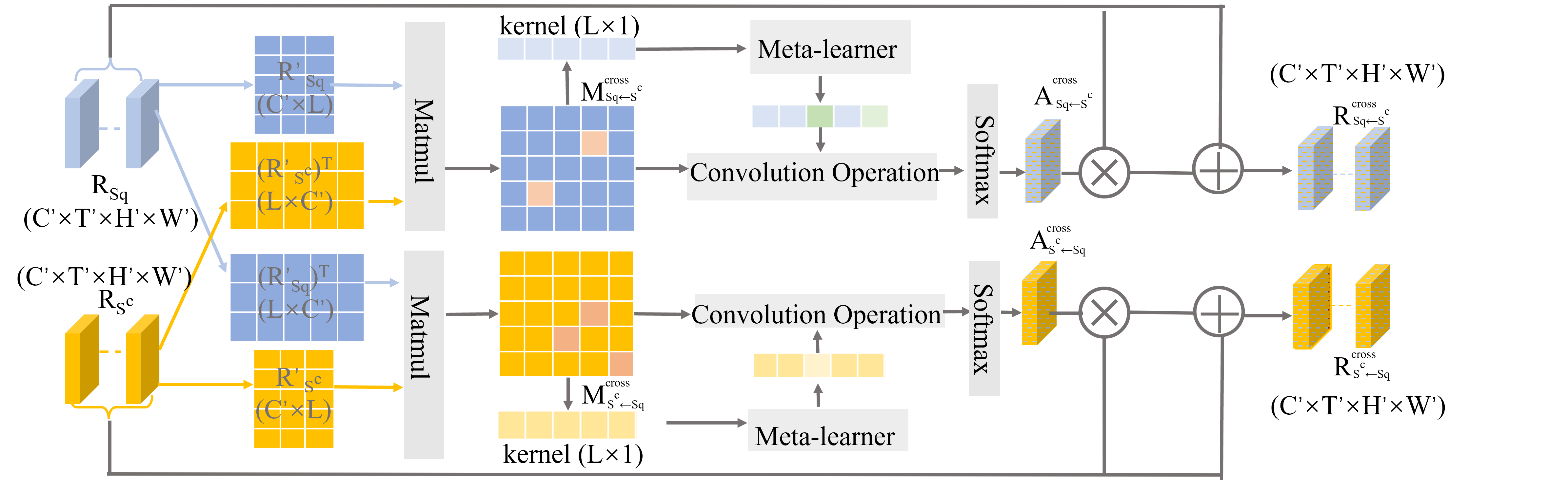}
\caption{Cross-attention module}
\label{cross-attention}
\end{figure*}

\noindent
\textbf{Cross-attention module}
While the self-attention module highlights the critical spatio-temporal region in the representation itself, the cross-attention module focuses on the correlation between the query video and the support set. As shown in Figure~\ref{cross-attention}, we follow the same steps as in the self-attention module to reshape each representation to $R_v^{'} \in \mathbb{R}^{C' \times L}$. After that, we compute the correlation map for each pair of the query video and the support class prototype. For example, for the pair of the query video $R_{S_q}$ and support class $c$,i.e., $R_{S^{c}}$, we compute the correlation map for the query video $M_{S_q \leftarrow S^{c}}^{cross}$ between the query video and support class:
\begin{equation}
M_{S_q \leftarrow S^{c}}^{cross} =(R'_{S^{c}})^{\intercal}R'_{S_q}.
\label{query-correlation-cross}
\end{equation}

Then for the support class c, the correlation map $M_{S^{c} \leftarrow S_q}^{cross}$ between the query video and support class is:
\begin{equation}
M_{S^{c} \leftarrow S_q}^{cross} =(R'_{S_q})^{\intercal}R'_{S^{c}}.
\label{cross-correlation-query}
\end{equation}
After getting the correlation map for query video and support class in each pair, we go through the same steps as in the self-attention module, which are shown in the Figure~\ref{cross-attention}, to get the cross-attention representation for query video and support class in each pair,i.e., $R_{S_q \leftarrow S^{c}}^{cross}$ and $R_{S^{c} \leftarrow S_q}^{cross}$.

\noindent
\textbf{Attention fusion module}
After we get the self-attention and cross-attention representation from the two attention modules, we compute the probability of predicting $S_q$ as the class $k$ using self-attention representation:
\begin{equation}
P_{self}(y=k|S_q) = \frac{exp(-D_{cos}(R_{S_q}^{self},R_{S^{k}}^{self}))}{\sum_{j=1}^{C}exp(-D_{cos}(R_{S_q}^{self},R_{S^{j}}^{self}))},
\label{distance-self-attention}
\end{equation}
where $D_{cos}$ denotes the cosine distance and $P_{self}(y=k|S_q)$ denotes the probability of predicting $S_q$ as the class $k \in \{1, 2,...,C\}$ using self-attention representations.
Then we compute the probability of predicting $S_q$ as the class $k$ using cross-attention representation:
\begin{equation}
P_{cross}(y=k|S_q) = \frac{exp(-D_{cos}(R_{S_q \leftarrow S^{k}}^{cross},R_{S^{k} \leftarrow S_q}^{cross}))}{\sum_{j=1}^{C}exp(-D_{cos}(R_{S_q \leftarrow S^{j}}^{cross},R_{S^{j} \leftarrow S_q}^{cross}))},
\label{distance-cross-attention}
\end{equation}
where $P_{cross}(y=k|S_q)$ denotes the probability of predicting $S_q$ as the class $k \in \{1, 2,...,C\}$ using cross-attention module.

To take advantage of the discriminative information from two attention mechanisms, we leverage the attention fusion module with the nearest neighbor classifier: 
\begin{equation}
P(y=k|S_q) = \frac{1}{2}[P_{self}(y=k|S_q) +P_{cross}(y=k|S_q)], 
\label{distance-fusion}
\end{equation}
where $P(y=k|S_q)$ denotes the final probability of predicting $S_q$ as the class $k \in \{1, 2,...,C\}$.

\noindent
\textbf{Multi-task training}
To reduce the risk of overfitting in the training dataset and generate a general representation for unseen class, we train the MASTAF model in a multi-task setting to regularize the embedding network. We combine the nearest neighbor classifier and the global video classifier.

During the training process, after the attention fusion module computes the probability of predicting query video to one of the classes in the support set, we use a negative log-probability as the loss function of the nearest neighbor classifier based on the actual class label:
\begin{equation}
L_1 = -\sum_{k=1}^{C}log P(y=k|S_q).
\label{L-1-LOSS}
\end{equation}

Since the representations after the cross-attention module contain highlighting regions related to the query video, we choose these representations to predict the global class in the whole training dataset. The total class number in the training dataset is $Z$. We feed these cross-attention representations to a fully connected layer and a softmax layer to get the probability of predicting the global class,i.e., $P(y=z|S^{c})$ where $z \in \{1, 2,...,Z\}$. Then we define the loss function of the global video classifier as:
\begin{equation}
L_2 = -\sum_{z=1}^{Z}log P(y=z|S^{c}).
\label{L-2-LOSS}
\end{equation}
Finally, the loss function of the MASTAF model is defined as:
\begin{equation}
L = L_1 + \lambda\  L_2,
\label{overall-LOSS}
\end{equation}
where we use $\lambda$ to weigh the impact of different classification tasks. Note that a multi-task training setting is only used during the training process. This setting is discarded at the inference stage.

\section{Evaluation}
\subsection{Experimental Setup}
\noindent
\textbf{Datasets.}
We compare MASTAF with existing work on UCF101~\cite{ucf101}, HMDB51~\cite{hmdb51}, and Something-Something V2 (SSv2)~\cite{ssv2}. We do not use Kinetics-100~\cite{cmn} to avoid bias because one of our MASTAF models is pre-trained on Kinetics-700~\cite{kinetics-700}. In these datasets, SSv2 is more challenging because it focuses on actions related to temporal relationships such as `pretending to take something from somewhere' versus `take something from somewhere'~\cite{reason}. There are two few-shot splits for SSv2 proposed by CMN~\cite{cmn} and OTAM~\cite{otam}, containing 64, 12, and 24 classes as the training, validation, and test set. We use SSv2-part and SSv2-all denote the split from CMN~\cite{cmn} and the split from OTAM~\cite{otam}. The difference between these two splits is the number of video samples in each class. For SSv2-part, Zhu and Yang~\cite{cmn} randomly selects 100 samples for each class, whereas for SSv2-all, Cao~\cite{otam} uses all the samples in the original SSv2. We evaluate our method in these two splits. Additionally, we also follow the split in ARN~\cite{arn} for HMDB51 and UCF101. 

\noindent
\textbf{Evaluation and baseline.}
Following the evaluation process in TRX~\cite{trx}, we evaluate the 5-way 1-shot and 5-way 5-shot video classification task and report the average accuracy over 10,000 randomly selected episodes from the test set. We compare our results with ten SOTA algorithms, i.e., TSN++~\cite{tsn}, CMN-J~\cite{cmn-j}, OTAM~\cite{otam}, FEAT~\cite{feat}, PAL~\cite{pal}, TRX~\cite{trx}, Baseline~\cite{close}, MatchingNet~\cite{matchingnet}, ProtoGAN~\cite{protogan}, ARN~\cite{arn}. In particular, the idea of Baseline (Notation from \cite{close}) is to train a new classifier with the given labeled examples in novel classes after extracting the representation from the embedding network. For a fair comparison, we use three MASTAF models with three different types of embedding networks, i.e., MASTAF-\{TSN\}, MASTAF-\{R3D\} and MASTAF-\{ViViT\}. For MASTAF-\{TSN\}, we follow the same embedding network configuration with \cite{otam,trx,feat,pal}, using an ImageNet pre-trained ResNet-50 as the backbone network. For MASTAF-\{R3D\}, we use the merged video dataset with Kinetics-700~\cite{kinetics-700},Moment-in-time~\cite{moment-in-times}, and START-action~\cite{stair} to pre-train 3D ResNet-50 embedding network. We also compare our approach against the previous work based on a 2D CNN embedding network where we replace 2D CNN with 3D CNN. We use Baseline-\{R3D\}, MatchingNet-\{R3D\}, TRX-\{R3D\} as the baselines by replacing the 2D CNN embedding network with a 3D CNN embedding network(same pre-trained R3D model as MASTAF-\{R3D\}). We extract one representation using pre-trained R3D from each video and then go through the matching part proposed in Baseline~\cite{close}, MatchingNet~\cite{matchingnet} and TRX~\cite{trx}. For MASTAF-\{ViViT\} and TRX-\{ViViT\}, we use ViViT~\cite{vivit} as our embedding network. We initialize ViViT from a ViT~\cite{vit} image model trained on the JFT~\cite{JFT} dataset. Due to the huge computation demand for ViViT~\cite{vivit}, we only perform 5-way 1-shot learning for MASTAF-\{ViViT\} and TRX-\{ViViT\}.

\noindent
\textbf{Experimental Configuration.} 
For MASTAF-\{TSN\}, MASTAF-\{ViViT\} and TRX-\{ViViT\}, we evenly sample 8 frames from each video as 8 segments for each video. For 3D CNN-based Baseline, MatchingNet, TRX and MASTAF-\{R3D\}, we evenly sample 16 frames from each video sample. After that,  we resize each frame to $256 \times 256$. Then we randomly flip each frame horizontally and crop the center region of $224 \times 224$ to augment the training data. For test data, we only crop the center with the same size without the horizontal flipping. Then for MASTAF-\{TSN\}, we use an ImageNet pre-trained ResNet-50 as the backbone and average all the frame representations as to the video representation. For 3D CNN-based Baseline, MatchingNet, TRX and MASTAF-\{R3D\}, we use a 3D ResNet-50\cite{would} with the weights pre-trained on the combined dataset with Kinetics-700\cite{kinetics-700}, Moments in Time\cite{moment-in-times}, and Start Action\cite{stair} as the embedding network. After finetuning in the validation dataset, we set 0.025 as the temperature hyperparameter($\tau$ in Eq~\ref{self-relation-softmax}) and set 6 as the meta-learner scaled dimension($l$ is the scaled dimension of $f_{\gamma}$ in Eq~\ref{meta-learner}), and set 2 as the loss weight hyperparameter($\lambda$ in Eq~\ref{overall-LOSS}). We train our model for 128,000 episodes in eight NVIDIA RTX A5000 GPU(except for the larger SSv2-all, we train our model for 256,000 episodes). We optimize the MASTAF model with SGD, in which the learning rate is 0.01. After fine-tuning, we adopt the batch-size of 128, 64, 32, 32 for UCF101, HMDB51, SSV2-part, and SSV2-all, respectively. 

\subsection{Comparison with State-of-the-art Algorithms}
\begin{figure}
\centering
\includegraphics[width=0.4\textwidth]{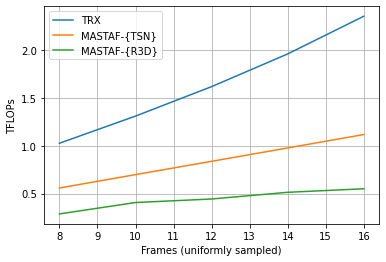}
\caption{Computational demand analysis for TRX, MASTAF-\{TSN\} and MASTAF-\{R3D\} as the number of sampled frames varies from 8 to 16 frames on UCF101}
\label{tflops}
\end{figure}

\begin{figure}
\centering
\includegraphics[width=0.4\textwidth]{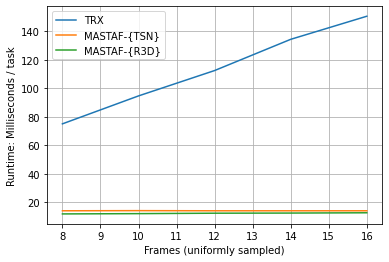}
\caption{Model inference's runtime analysis for TRX, MASTAF-\{TSN\} and MASTAF-\{R3D\} in one NVIDIA RTX A5000 GPU as the number of sampled frames varies from 8 to 16 frames on UCF1011}
\label{runtime}
\end{figure}

Table~\ref{start-of-art-ssv2} tabulates the overall 5-way 1-shot and 5-way 5-shot performance compared with existing methods on two splits of SSv2. We can categorize these comparative methods into three groups based on the embedding network. In 2D CNN embedding group, TSN++~\cite{tsn}, CMN-J~\cite{cmn-j}, FEAT~\cite{feat} are model agnostic and do not apply any frame-level temporal alignment. Compared with these three methods, the other three methods,i.e., OTAM~\cite{otam}, PAL~\cite{pal}, and TRX~\cite{trx}, adopt frame-level temporal alignment, which further improves the performance of few-shot video classification. MASTAF-\{TSN\} outperforms existing 5-way 1-shot video classification algorithms in the 2D CNN group. TRX~\cite{trx} achieves SOTA performance for 5-way 5-shot learning because it leverages the temporal information from different frames in different videos in the support set. However, this complex alignment strategy leads to huge computation costs and increases model inference's runtime. Figure~\ref{tflops} and Figure~\ref{runtime} compare the TFLOPs and model inference's runtime of TRX~\cite{trx} and MASTAF-\{TSN\}. Our approach achieves SOTA accuracy without increased computational cost and is more efficient than TRX~\cite{trx}. As the number of frames sampled from a video increases, TRX~\cite{trx} consumes more computational resources and takes longer for the inference process.
\begin{table*}[h]
\begin{center}
\caption{Comparison on 5-way 1-shot and 5-shot benchmarks of SSv2-part, and SSv2-all. The best performance in each group is highlighted. $\dagger$: Results from \cite{otam}. *: Results from \cite{pal} }
\label{start-of-art-ssv2}
\begin{tabular}{lccccc}
\hline
\multirow{2}{*}{Method} & \multirow{2}{*}{\begin{tabular}[c]{@{}c@{}}Embedding\\ Groups\end{tabular}} & \multicolumn{2}{c}{SSv2-part} & \multicolumn{2}{c}{SSv2-all}  \\ \cline{3-6} 
                        &                                                                             & 1-shot        & 5-shot        & 1-shot        & 5-shot        \\ \hline
TSN$++^{\dagger}$~\cite{tsn}                  & \multirow{7}{*}{2D CNN}                                                     & -             & -             & 34.4          & 43.8          \\
CMN-J~\cite{cmn-j}                   &                                                                             & 36.2          & 48.8          & -             & -             \\
FEAT*~\cite{feat}                    &                                                                             & -             & -             & 45.3          & 61.2          \\
OTAM~\cite{otam}                    &                                                                             & -             & -             & 42.8          & 52.3          \\
PAL~\cite{pal}                     &                                                                             & -             & -             & 46.4          & 62.6          \\
TRX~\cite{trx}                     &                                                                             & 36.0          & \textbf{59.1} & 42.0          & \textbf{64.6} \\
MASTAF-\{TSN\}          &                                                                             & \textbf{37.5} & 50.2          & \textbf{46.9} & 62.4          \\ \hline
Baseline-\{R3D\}             & \multirow{4}{*}{3D CNN}                                                     & 24.9           & 36.1           & 25.6           & 39.8 \\
MatchingNet-\{R3D\}             &                                                     & 34.1           & 45.2           & 43.2           & 54.4 \\
TRX-\{R3D\}             &                                                      & 26.1           & 47.0           & 34.9           & 58.9           \\
MASTAF-\{R3D\}          &                                                                             & \textbf{39.9} & \textbf{52.2} & \textbf{50.3} & \textbf{66.7} \\ \hline
TRX-\{ViViT\}           & \multirow{2}{*}{Transformer}                                                & 34.7           & -             & 42.7           & -             \\
MASTAF-\{ViViT\}        &                                                                             & \textbf{45.6} & -             & \textbf{60.7} & -             \\ \hline
\end{tabular}
\end{center}
\end{table*}
In the 3D CNN group, the accuracy of the TRX-\{R3D\} is lower than TRX because it cannot perform the frame-level temporal alignment. For PAL~\cite{pal}, Zhu~\cite{pal} also mentioned that 3D CNN models~\cite{I3D,c3d,S3D-G} do not perform better than 2D CNN models due to the lacking of frame-level similarity scores. In comparison, MASTAF-\{R3D\} takes advantage of the spatio-temporal representation from R3D and further improves the performance. In the Transformer group, MASTAF-\{ViViT\} further enhances the performance. These results demonstrate MASTAF works best when spatio-temporal information is well represented in advanced video classification models. In contrast, existing work with a 2D embedding network cannot maintain high performance when replacing a 2D CNN embedding network with other advanced video representation models. 

Table~\ref{start-of-art-ucf-hmdb} tabulates the overall 5-way 1-shot and 5-way 5-shot performance compared with existing methods on UCF101 and HMDB51. Our MASTAF with a 2D embedding network achieves decent performance while TRX and PAL achieve SOTA accuracy on these two datasets. The reason is that TSN does not provide enough spatio-temporal information for MASTAF to distinguish the query video from the videos in the support set. So to benefit the most from MASTAF, we explore our MASTAF with a 3D CNN embedding network and Video Transformer. As shown in Table~\ref{start-of-art-ucf-hmdb}, our MASTAF-\{R3D\} outperforms other methods based on 3D models and MASTAF-\{ViViT\} outperforms TRX-\{ViViT\} and achieves new SOTA performance. Compared with MASTAF-\{TSN\}, MASTAF-\{R3D\} has significantly lower resource consumption and running time, as shown in Figure~\ref{tflops} and Figure~\ref{runtime}. 
\setlength{\tabcolsep}{4pt}
\begin{table*}[h]
\begin{center}
\caption{Comparison on 5-way 1-shot and 5-shot benchmarks of UCF101, and HMDB51. The best performance in each group is highlighted. *: Results from \cite{pal}}
\label{start-of-art-ucf-hmdb}
\begin{tabular}{lccccc}
\hline
\multirow{2}{*}{Method} & \multirow{2}{*}{\begin{tabular}[c]{@{}c@{}}Embedding\\ Groups\end{tabular}} & \multicolumn{2}{c}{UCF101} & \multicolumn{2}{c}{HMDB51} \\ \cline{3-6} 
                        &                                                                             & 1-shot   & 5-shot          & 1-shot   & 5-shot          \\ \hline
FEAT*~\cite{feat}                    & \multirow{4}{*}{2D CNN}                                                    & 83.9     & 94.5            & 60.4     & 75.2            \\
PAL~\cite{pal}                     &                                                                             & \textbf{85.3}     & 95.2            & \textbf{60.9}     & \textbf{75.8}            \\
TRX~\cite{trx}                     &                                                                             & -        & \textbf{96.1}            & -        & 75.6            \\
MASTAF-\{TSN\}           &                                                                             & 79.3        & 90.3               & 54.8        & 67.7               \\ \hline
ProtoGAN~\cite{protogan}                & \multirow{6}{*}{3D CNN}                                                    & 57.8     & 80.2            & 34.7     & 54              \\
ARN~\cite{arn}                     &                                                                             & 66.3     & 83.1            & 45.5     & 60.6            \\
Baseline-\{R3D\}             &                                                                             & 53.4        & 88.7               & 40.1        & 68.1               \\
MatchingNet-\{R3D\}             &                                                                             & 82.7        & 93.5               & 61.8        & 75.6               \\
TRX-\{R3D\}             &                                                                             & 82.5        & 94.1               & 57.0        & 74.3               \\
MASTAF-\{R3D\}            &                                                                             & \textbf{90.6}     & \textbf{97.6}   & \textbf{67.9}     & \textbf{81.2}   \\ \hline
TRX-\{ViViT\}           & \multirow{2}{*}{Transformer}                                                & 84.8           & -             & 58.1           & -             \\
MASTAF-\{ViViT\}        &                                                                             & \textbf{91.6} & -             & \textbf{69.5} & -             \\ \hline
\end{tabular}
\end{center}
\end{table*}
\subsection{Ablation study}
We have shown in Section 4.2 that our MASTAF can make the most of the advanced video classification model to improve the accuracy without more computational cost. We now perform detailed ablation studies on two dataset UCF101 and SSV2-all to show each module's influence. The ablation studies about multi-task learning setting, meta-learner and  residual structure are in the Appendix. In these ablation studies, all MASTAF models use the 3D ResNet-50 model pre-trained on the merged video dataset with Kinetics-700~\cite{kinetics-700},Moment-in-time~\cite{moment-in-times}, and START-action~\cite{stair} as the embedding network.

\begin{table}
\begin{center}
\caption{Comparison results with three variants of MASTAF for 5-way 1-shot video classification}
\label{fusion_results}
\begin{tabular}{lcc}
\hline
Method     & \multicolumn{1}{l}{UCF101} & \multicolumn{1}{l}{SSv2-all} \\ \hline
MASTAF-Neighbor    & 82.7                       & 43.2                               \\ 
MASTAF-Self  & 90.3                       & 49.4                               \\ 
MASTAF-Cross & 90.5                       & 49.2                               \\ 
MASTAF       & \textbf{90.6}              & \textbf{50.3}                      \\ \hline
\end{tabular}
\end{center}
\end{table}
\subsubsection{Attention fusion mechanism}
\label{sss:attention}
To explore the effectiveness of the attention fusion mechanism, we introduce three comparison models, i.e., MASTAF-Neighbor, MASTAF-Self, MASTAF-Cross. In MASTAF-Neighbor, representations learned from the embedding network are fed into the nearest neighbor classifier and a global video classifier directly without our attention mechanisms. For MASTAF-Self and MASTAF-Cross, before being fed into two classifiers, representations go through the self-attention and cross-attention mechanism, respectively. Table~\ref{fusion_results} shows the comparison results. 
Compared with MASTAF-Neighbor, after adding the attention mechanism, all three other models have a significant performance improvement, demonstrating that representations after the embedding network have some spatio-temporal features related to the non-target action region. The cross-attention mechanism in MASTAF-Cross aid in highlighting the spatio-temporal features associated with the target action region among the query video and support set. MASTAF-Self's self-attention module helps highlight spatio-temporal features related to the action in each video itself. Therefore, combining two different attention modules can take advantage of each module to further extract more discriminative spatio-temporal representations. The results in Table~\ref{fusion_results} demonstrate our argument. 

\section{Conclusion}
This paper proposes a Model-Agnostic Spatio-Temporal Attention Fusion network(MASTAF) for few-shot video classification. MASTAF is a simple and efficient few-shot video classification framework compatible with different video classification models. MASTAF make the most of the knowledge learned from the advanced video classification model and uses self- and cross-attention to highlight the spatio-temporal features. MASTAF works best when spatio-temporal information is well represented in advanced video classification models and improves the state-of-the-art performance of 5-way 1-shot, and 5-shot video classification on UCF101, HMDB51, and SSv2, e.g., MASTAF improves the accuracy of 5-way 1-shot video classification to 91.6\%, 69.5\%, and 60.7\% for UCF101, HMDB51, and SSv2, respectively.

{\small
\bibliographystyle{ieee_fullname}
\bibliography{egbib}
}

\end{document}